\documentclass[11pt]{article}
\usepackage[preprint]{acl}
\usepackage{times}
\usepackage{latexsym}
\usepackage[T1]{fontenc}
\usepackage[utf8]{inputenc}
\usepackage{microtype}
\usepackage{inconsolata}
\usepackage{graphicx}
\usepackage{booktabs}
\usepackage{amsmath}
\usepackage{amsfonts}
\usepackage{amssymb}
\usepackage{amsthm}
\usepackage{tikz}
\usetikzlibrary{positioning, arrows.meta, calc}
\tikzstyle{process} = [rectangle, rounded corners, minimum width=3.5cm, minimum height=1cm, text centered, draw=black, font=\sffamily\bfseries]
\tikzstyle{arrow} = [thick,->,>=Stealth]

\title{CheckMIABench: Firm Foundations For Membership Inference Attacks on Language Models}

\author{
 \textbf{Jeffrey G. Wang\textsuperscript{$\ast$1}},
 \textbf{Jason Wang\textsuperscript{$\ast$1}},
 \textbf{Marvin Li\textsuperscript{$\ast$1}},
 \textbf{Seth Neel\textsuperscript{2}}
\\
 \textsuperscript{1} Harvard University,
 \textsuperscript{2} Harvard Business School
\\
\small{\textbf{Correspondence}: \texttt{\{jgwang21, sethneel.ai\}@gmail.com}}
}

\begin{document}
\maketitle
\renewcommand{\thefootnote}{\fnsymbol{footnote}}
\footnotetext[1]{Equal contribution.}
\renewcommand{\thefootnote}{\arabic{footnote}}
\begin{abstract}
Membership inference attacks (MIAs) are a canonical way to assess a machine learning model's privacy properties. Although several attempts have been made to evaluate MIAs on language models, the extant literature has suffered numerous difficulties in constructing clean evaluations to test new techniques. In particular, subtle distribution shifts between member and non-member sets can undermine the statistical validity of MIAs; recent work has underscored this by showing that ``blind'' methods with no access to the underlying model can perform far better than published methods on the same benchmarks. This paper constructs a benchmark for principled evaluation of MIAs against LLMs, by leveraging the insight that training data before and after a fixed point during training are drawn from the same distribution. Therefore, all open-source models with intermediate checkpoints and public training data can be converted into MIA testbeds. We apply our framework to a half-dozen published attacks on the Pythia and OLMo family of models, from 70M to 7B parameters. To facilitate further privacy research, we open-source a modular library for designing and implementing attacks in this setting: \url{https://github.com/safr-ai-lab/pandora_llm}.
\end{abstract}

\section{Introduction}
Large language models (LLMs) have become indispensable workhorses for knowledge-intensive tasks, from summarizing medical records and drafting clinical notes to screening legal contracts and flagging anomalous financial transactions. As these models are increasingly deployed in high-stakes arenas such as healthcare and finance, privacy has become a first-order requirement for responsible use. As such, recent security research has revisited a threat model long studied for classifiers and regressors: that of membership inference attacks (MIAs) \citep{dl_privacy_survey, shokri2017membership}. In an MIA, an adversary with some level of access to the model attempts to decide whether a particular example was used in a model's training data (``member") or is merely drawn i.i.d. from the same distribution (``non-member"). Highly accurate MIAs against LLMs would be useful not only to demonstrate privacy leakage, but could also be adopted as practical probes for related phenomena, including quantifying memorization during training \citep{zhou2023quantifying} and empirically evaluating the success of machine unlearning techniques \citep{kurmanji2023unbounded, pawelczyk2023incontext, hayes2024inexact}. 

While there has been a recent surge of research on MIAs against LLMs \citep{duan2024membership, li-etal-2023-mope, mattern2023membership}, it has been notoriously difficult to implement a correct MIA evaluation benchmark for pre-trained LLMs~\citep{das2025blindbaselinesbeatmembership, maini2024reassessingemnlp2024s}. In particular,~\citet{das2025blindbaselinesbeatmembership} showed that many evaluations for MIAs commonly used in the literature are beaten by simple supervised learning methods trained on i.i.d splits of member/non-member data that have no access to the underlying model. This means the accuracy of these MIAs primarily emerges from the distributional differences between member and non-member examples rather than the privacy properties of underlying LLM. For instance,~\citet{shi2023detecting} proposed using \emph{temporal differences} (from training data cutoffs of models) to construct member and non-member sets with the WikiMIA benchmark, where Wikipedia articles written before Jan 1st, 2017 were treated as the member data and articles after Jan 1st, 2023 were treated as non-member data. While reasonable at first glance, as such cutoffs enforce the member/non-member distinction, these sets are separable for another reason: their contents are distinct due to changing writing patterns over time. Indeed, ~\citet{das2025blindbaselinesbeatmembership} trained a simple bag-of-words classifier that could distinguish between member and non-member data without the underlying model with a True Positive Rate (TPR) of $94.7$\% at a False Positive Rate (FPR) at $5$\%, far outperforming the best attack proposed on the benchmark. 


In this paper, we begin with an overview of the challenges researchers have faced in creating theoretically clean evaluations for membership inference. Then, we instantiate a cleaner dataset split for two common open LLM families, sanity checking them with ``blind baselines" to verify the splits do not suffer from those same challenges. Finally, we benchmark a slate of common methods on these clean splits, finding more limited performance.

\section{Related Work}
MIAs refer to a class of methods that can determine if a given data point $z$ was included in the training dataset of a particular model $\theta$. They were initially motivated by privacy considerations; if an adversary can determine if $z$ was used to train $\theta$ with accuracy higher than the base rate, then information theoretically $\theta$ \emph{must} encode some information about $z$ that the attack is able to leverage. The setup to evaluate such MIAs is fairly straightforward: given an initial dataset $D$, some subset of the data is set aside prior to training (or sampled from the distribution after training) as a validation set $D_{\rm {nonmem}}$, and some dataset $D_{\rm {mem}}$ is used for training to produce a new model $\theta$. Then the following process is repeated a set number of times:
\begin{enumerate}
    \item Flip a fair coin. If heads, sample with replacement $z \sim D_{\rm {mem}}$; otherwise, sample $z \sim D_{\rm {nonmem}}$. 
    \item Given $z$ and some access to $\theta$, the MIA produces a score $p$ for that data point. 
\end{enumerate}
Given a membership inference score, for any threshold $\tau$, there is a corresponding membership inference attack that predicts $z \in D_{\rm mem}$ if and only if $p < \tau$. Each $\tau$ corresponds to a point on the Receiver Operating Characteristic (ROC) curve for the binary classification. As is common in the literature, to evaluate MIAs in this paper, we report the area under the ROC curve (AUC) as well as the TPR at low FPR of each method. The latter metric is widely used in the literature because any attack which can extract with high confidence even a small fraction of the training data poses a serious privacy risk~\citep{carlini2022first}. 


One key assumption underlies this evaluation scheme: individual samples in the training and validation data are drawn i.i.d. from the same distribution $\mathcal{P}.$ Clearly, if $\theta$ contains no information about $D_{\rm {mem}}$ then the maximum attack accuracy is $50\%$, since $D_{\rm {mem}}$  and $D_{\rm {nonmem}}$ have identical marginal distributions. Any additional information about whether $z \sim D_{\rm {mem}}$ or $z \sim D_{\rm {nonmem}}$ must come from the model $\theta$. Indeed, if $D_{\rm {mem}}$ and $D_{\rm {nonmem}}$ differ in some way that is independent of whether $\theta$ is trained on it, then the maximum attack accuracy can be $100\%$ even if the $\theta$ contains no information about $D_{\rm {mem}}$. For instance, if one took a standard training and validation dataset for an LLM and prepended \texttt{The quick, brown fox jumps} to only the train dataset, an MIA without any access to $\theta$ could still achieve near-perfect accuracy by simply detecting if a given $z$ begins with this phrase. 

\paragraph{Existing Evaluations.} Existing evaluations for MIAs fall in three broad categories. Several papers introduce benchmarks using the temporal cutoff approach of WikiMIA \cite{shi2023detecting, liu2024probinglanguagemodelspretraining}. Others utilize train/val splits of open models, most commonly using The Pile dataset with the Pythia family of models \cite{biderman2023pythia, gao2020pile}. The MIMIR benchmark refines this approach, with an additional deduplication step of the non-member sets against training data\footnote{They create several member/non-member splits such that all non-member points with more than a $p$-proportion overlap in $n$-grams with any training data point are removed, for various settings of $n$ and $p$.} \cite{duan2024membership}. 



\paragraph{Challenges of Deduplication.} The Pythia family of models includes those trained on a deduplicated version of The Pile's training set; however, since only the training data is deduplicated, it has a different distribution than the validation split of The Pile. One solution, used in MIMIR, deduplicates the validation set against the train set---but this approach is not perfectly sound either, as this induces a different marginal distribution on member vs. non-member points. To see this, consider the setting where there are two very rare documents in the corpus. If both documents are split into the training corpus, only one will be included post deduplication; on the other hand, if they are both included in the validation split, they will both make it through pre-processing since they will only be deduplicated against the training set. Indeed, ~\citet{meeus2025sokmembershipinferenceattacks} find that a bag-of-words classifier is able to achieve extremely high AUCs on certain splits (up to $0.86$), implying that deduplication causes distribution shifts that violate the MIA assumptions. The correct approach then, would jointly deduplicate the entire corpus prior to randomly splitting it into train and validation sets, which can't be done ex-post. Joint deduplication would also mitigate the issue of ``fuzzy membership'' that also poses a barrier to rigorous MIA evaluations in LLMs: ~\citet{duan2024membership} show there is substantial overlap between the training and validation sets for The Pile, which is the dataset used to train the Pythia and GPT-NeoX series of models, an oft-used benchmark for MIA papers. 




It is important to note that these difficulties are in some sense unique to LLMs; in classical supervised learning, the evaluator will typically have access to the entire dataset and can choose the train/validation partition before training the model.

\paragraph{Concurrent Work.} ~\citet{kim2025detectingtrainingdatalarge} also suggest a checkpoint-based approach to evaluate MIAs on OLMo, but they neither verify that their member/non-member splits are clean via blind baselines nor release a checkpoint-based MIA dataset for The Pile, as we do in this paper. While our checkpoint-based benchmark demonstrates that existing MIAs fail to achieve statistically-significant results, ~\citet{hayes2026exploringlimitsstrongmembership} more recently demonstrated that scaling up strong white-box MIAs can succeed for LLMs even in the pre-training setting. This method requires the pre-training of many reference models and is thus extremely computationally expensive, so we are unable to evaluate it on our benchmark.
\section{Our MIA Evaluation Pipeline}
\label{sec:pipeline}
In this section, we will describe a clean method to derive member and non-member splits. This method only relies on access to the model checkpoints in the middle of training and the training data order, which is available for many models, like the Pythia family or OLMo \cite{groeneveld2024olmoacceleratingsciencelanguage,kim2025detectingtrainingdatalarge}. We will also describe the procedure to generate member and non-member data using different checkpoints during the training, checking it against blind MIAs as in~\citep{das2025blindbaselinesbeatmembership,maini2024reassessingemnlp2024s}.



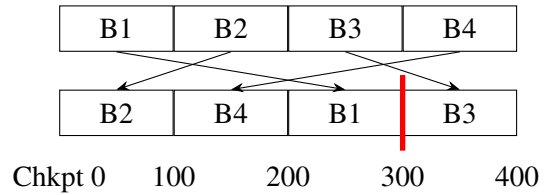
\begin{figure}[h!]
\centering
\begin{tikzpicture}[
  box/.style={draw, minimum width=1.5cm, minimum height=0.6cm, align=center},
  >=Stealth
]

\node[box] (t1) at (0,0) {B1};
\node[box, right=0pt of t1] (t2) {B2};
\node[box, right=0pt of t2] (t3) {B3};
\node[box, right=0pt of t3] (t4) {B4};

\node[box, below=0.5cm of t1] (b1) {B2};
\node[box, right=0pt of b1] (b2) {B4};
\node[box, right=0pt of b2] (b3) {B1};
\node[box, right=0pt of b3] (b4) {B3};
\node[below=1.5em of $(b1.west)$] {Chkpt 0};
\node[below=1.5em of $(b1.east)!0.5!(b2.west)$] {100};
\node[below=1.5em of $(b2.east)!0.5!(b3.west)$] {200};
\node[below=1.5em of $(b3.east)!0.5!(b4.west)$] {300};
\node[below=1.5em of $(b4.east)$] {400};

\draw[->] (t1.south) -- (b3.north);
\draw[->] (t2.south) -- (b1.north);
\draw[->] (t3.south) -- (b4.north);
\draw[->] (t4.south) -- (b2.north);
\draw[red, line width=2pt] ([yshift=0.5cm]b3.east) -- ([yshift=-0.5cm]b3.east);

\end{tikzpicture}
\caption{The shuffling of packed samples to construct the training data of Pythia and our checkpoint-based method to evaluate MIAs. In this scenario, we would sample from chunks \{B2,B4,B1\} for member data and chunk \{B3\} for the non-member data to evaluate MIAs for the model at Checkpoint $300$.}
\label{fig:schematic_shuffle}
\end{figure}

\paragraph{Member and Non-member Data Generation.} Recall that the LLM data collection and training procedure typically involves forming documents taken from links scraped from the web, which are then tokenized and packed into training examples of a fixed size. These packed examples are deduplicated against each other and shuffled to form the \emph{training order} of the model (Figure~\ref{fig:schematic_shuffle}). This shuffling of the training order will be the key to our MIA evaluation setup. In our setup, we select a model, e.g., Checkpoint $300$ in Figure~\ref{fig:schematic_shuffle}, using the data before the checkpoint as the member data (herein \{B2,B4,B1\}) and the data after the checkpoint (herein \{B3\}) as the non-member data. We then evaluate the model at that checkpoint. Because the data before and after are drawn from the same distribution, the member and non-member data have the same marginal distributions.

\paragraph{Generating and Learning from Features.} Using member and non-member data generated in the previous steps, we can evaluate different MIAs. We release our codebase as a modular Python library, with documentation and tests, that allows one to implement and benchmark new MIAs in this setting. We also implement (and in Section~\ref{sec:results}, benchmark) many MIAs from extant literature, including: simple loss thresholding~\citep{yeom2018privacy}, Min-K~\citep{shi2023detecting}, Min-K\%++~\citep{zhang2025minkimprovedbaselinedetecting}, Zlib entropy ~\citep{carlini2021extracting}, ReCaLL ~\citep{xie2024recallmembershipinferencerelative}, and MoPe~\citep{li-etal-2023-mope}. 

\paragraph{Blind MIAs.} Finally, our pipeline implements many ``blind" MIAs as a baseline check that member and non-member data don't have distributional differences easily checkable with a simple supervised learning method. For a given set of splits, we train classifiers on statistical representations of the text, including Bag of Words, TF-IDF, Word2Vec representations~\citep{mikolov2013efficientestimationwordrepresentations}, and BERT embeddings~\citep{devlin2019bertpretrainingdeepbidirectional}. Full details on the training details of these classifiers are available in the Appendix. 


\section{Results}
\label{sec:results}

\paragraph{Setup.} We instantiate our evaluation on the Pythia family of models, which were trained for $143,\!000$ optimizer steps on a deduplicated version of The Pile, representing approximately $1.5$ epochs. We evaluate the model at step $97,\!000$ (over $95$\% of the way through a full epoch), uniformly sampling data points before and after this step.\footnote{Note that the use of intermediate checkpoints is an additional access assumption, but most literature already assumes access to a $1$-epoch intermediate checkpoint to ensure that each datapoint is seen a uniform number of times.}

We also benchmark our method on the OLMo family of models trained on the Dolma dataset~\citep{groeneveld2024olmoacceleratingsciencelanguage,dolma}. Due to computational constraints, we only benchmark OLMo results on the $7$B parameter model. We use the checkpoint at step $400,\!000$, which represents $88$\% of the way through the first epoch of training. See full results in Appendix \ref{apdx:olmo}.

\paragraph{Blind MIAs.} First, we validate that blind MIAs without any access to the model, like supervised learning techniques on a split of member and non-member data points, get no signal in our setting (Table \ref{table:baselines}). In all settings, we train three different classifiers (logistic regression, random forest, and a neural network) on $4,\!000$ points from member and non-member classes using the tokens or text alone; then we evaluate our classifier on $1,\!000$ points from each class. We report the best test AUC among these three methods in Tables~\ref{table:baselines} and~\ref{tab:olmobaselines}, providing standard errors from $1{,}000$ bootstraps. Full details on the supervised classifiers are in Appendix \ref{apdx:supclas}. 

\begin{table}[h!]
\centering
\caption{Sanity check: supervised model-free (``blind") MIAs get no signal on our cleaned split of The Pile.}
\label{table:baselines}
\begin{tabular}{rcccc}
\toprule
\textbf{Blind MIA} & \textbf{AUC} & \textbf{AUC SE} & $\text{\textbf{TPR}}_{1\%}$ \\ \midrule
BoW (Tokens) & 0.505 & 0.0128 & 0.01 \\
TFIDF (Tokens) & 0.503 & 0.0127 & 0.006 \\
W2V (Text) & 0.490 & 0.0131 & 0.015 \\
BERT (Text) & 0.497 & 0.0127 & 0.012 \\
\bottomrule
\end{tabular}
\end{table}

\paragraph{Existing MIAs.}

After validating our dataset, we benchmark the MIAs listed in Section~\ref{sec:pipeline}, which we implement in our library. We find that current MIAs have limited success on our dataset drawn from the pre-training data (see Table~\ref{tab:results}). Note that this does not exclude the possibility that on certain subsets of the training data (in particular, the GitHub documents in The Pile), MIAs can still be performant, which was found in previous works~\citep{duan2024membership}. We note that while one attack, MoPe, achieves performance that is $2.5$x above the random baseline as measured by $\text{\textbf{TPR}}_{1\%}$ on the $70$m model, this performance diminishes on larger models. None of the attacks proposed achieve AUCs that are statistically significantly above the blind baseline of $0.5$ across any model size from $70$m to $2.8$b.


\begin{table}[h!]
\centering
\caption{We benchmark several MIAs against the Pythia family of models on our cleaned split of The Pile.}
\label{tab:results}
\begin{tabular}{rcccc}
\toprule
\textbf{Model} & \textbf{MIA} & \textbf{AUC} & \textbf{AUC SE} & $\text{\textbf{TPR}}_{1\%}$ \\ \midrule


70m & LOSS & 0.513 & 0.0130 & 0.006 \\
70m & Min-K & 0.500 & 0.0130 & 0.012 \\
70m & Min-K++ & 0.496 & 0.0128 & 0.011 \\
70m & Zlib & 0.499 & 0.0129 & 0.016 \\
70m & MoPe & 0.487 & 0.0126 & 0.025 \\
70m & ReCaLL & 0.490 & 0.0130 & 0.010 \\


160m & LOSS & 0.513 & 0.0130 & 0.007 \\
160m & Min-K & 0.500 & 0.0127 & 0.011 \\
160m & Min-K++ & 0.499 & 0.0126 & 0.014 \\
160m & Zlib & 0.499 & 0.0127 & 0.016 \\
160m & MoPe & 0.495 & 0.0132 & 0.015 \\
160m & ReCaLL & 0.490 & 0.0124 & 0.006 \\


410m & LOSS & 0.512 & 0.0128 & 0.007 \\
410m & Min-K & 0.498 & 0.0132 & 0.006 \\
410m & Min-K++ & 0.496 & 0.0126 & 0.014 \\
410m & Zlib & 0.499 & 0.0127 & 0.016 \\
410m & MoPe & 0.504 & 0.0130 & 0.014 \\
410m & ReCaLL & 0.493 & 0.0131 & 0.008 \\

1b & LOSS & 0.513 & 0.0122 & 0.007 \\
1b & Min-K & 0.502 & 0.0125 & 0.007 \\
1b & Min-K++ & 0.498 & 0.0126 & 0.013 \\
1b & Zlib & 0.500 & 0.0129 & 0.016 \\
1b & MoPe & 0.495 & 0.0126 & 0.015 \\
1b & ReCaLL & 0.487 & 0.0123 & 0.008 \\



2.8b & LOSS & 0.506 & 0.0131 & 0.007 \\
2.8b & Min-K & 0.489 & 0.0124 & 0.006 \\
2.8b & Min-K++ & 0.489 & 0.0126 & 0.014 \\
2.8b & Zlib & 0.499 & 0.0129 & 0.016 \\
2.8b & MoPe & 0.493 & 0.0131 & 0.006 \\
2.8b & ReCaLL & 0.498 & 0.0131 & 0.010 \\

\bottomrule
\end{tabular}
\end{table}
\section{Conclusion}

In this paper, we identify subtle distributional pitfalls of previous MIA evaluations for LLMs and propose a principled framework to avoid them. Our pipeline supports several popular open model families and we benchmark many existing attacks on it. Finally, this work provides a clear call-to-action for future model releases: the underlying dataset should be processed and deduplicated in tandem, then split into training and validation sets to ensure matching training and validation distributions.

\section{Limitations}

While our setting is general and can be instantiated across many open model families, there are several limitations and areas for future work. First, our framework requires open training data, model checkpoints, and the exact permutation of training examples throughout training. While many open model families provide this information, others, e.g., Llama and Qwen, do not~\cite{touvron2023llama, deepseekai2025deepseekv3technicalreport, yang2025qwen3technicalreport}. Despite this important limitation to any checkpoint-based approach, we argue that this methodology enables us to investigate MIA efficacy much more rigorously than existing work, and hope that it encourages model providers to release this auxiliary information in the future. Second, due to computational and data constraints, we only evaluated MIAs on models pretrained almost exclusively on English web data. Finally, in our experimental setup, we do not benchmark any models larger than $7$ billion parameters, again due to compute constraints. We are excited by future work verifying the extent of our negative results across scales, data subsets, languages, and training paradigms.








\bibliography{bibliography}

@misc{yang2025qwen3technicalreport,
      title={Qwen3 Technical Report}, 
      author={An Yang and Anfeng Li and Baosong Yang and Beichen Zhang and Binyuan Hui and Bo Zheng and Bowen Yu and Chang Gao and Chengen Huang and Chenxu Lv and Chujie Zheng and Dayiheng Liu and Fan Zhou and Fei Huang and Feng Hu and Hao Ge and Haoran Wei and Huan Lin and Jialong Tang and Jian Yang and Jianhong Tu and Jianwei Zhang and Jianxin Yang and Jiaxi Yang and Jing Zhou and Jingren Zhou and Junyang Lin and Kai Dang and Keqin Bao and Kexin Yang and Le Yu and Lianghao Deng and Mei Li and Mingfeng Xue and Mingze Li and Pei Zhang and Peng Wang and Qin Zhu and Rui Men and Ruize Gao and Shixuan Liu and Shuang Luo and Tianhao Li and Tianyi Tang and Wenbiao Yin and Xingzhang Ren and Xinyu Wang and Xinyu Zhang and Xuancheng Ren and Yang Fan and Yang Su and Yichang Zhang and Yinger Zhang and Yu Wan and Yuqiong Liu and Zekun Wang and Zeyu Cui and Zhenru Zhang and Zhipeng Zhou and Zihan Qiu},
      year={2025},
      eprint={2505.09388},
      archivePrefix={arXiv},
      primaryClass={cs.CL},
      url={https://arxiv.org/abs/2505.09388}, 
}

@misc{xie2024recallmembershipinferencerelative,
      title={{ReCaLL}: Membership Inference via Relative Conditional Log-Likelihoods}, 
      author={Roy Xie and Junlin Wang and Ruomin Huang and Minxing Zhang and Rong Ge and Jian Pei and Neil Zhenqiang Gong and Bhuwan Dhingra},
      year={2024},
      eprint={2406.15968},
      archivePrefix={arXiv},
      primaryClass={cs.CL},
      url={https://arxiv.org/abs/2406.15968}, 
}

@misc{deepseekai2025deepseekv3technicalreport,
      title={{DeepSeek-V3} Technical Report}, 
      author={DeepSeek-AI and Aixin Liu and Bei Feng and Bing Xue and Bingxuan Wang and Bochao Wu and Chengda Lu and Chenggang Zhao and Chengqi Deng and Chenyu Zhang and Chong Ruan and Damai Dai and Daya Guo and Dejian Yang and Deli Chen and Dongjie Ji and Erhang Li and Fangyun Lin and Fucong Dai and Fuli Luo and Guangbo Hao and Guanting Chen and Guowei Li and H. Zhang and Han Bao and Hanwei Xu and Haocheng Wang and Haowei Zhang and Honghui Ding and Huajian Xin and Huazuo Gao and Hui Li and Hui Qu and J. L. Cai and Jian Liang and Jianzhong Guo and Jiaqi Ni and Jiashi Li and Jiawei Wang and Jin Chen and Jingchang Chen and Jingyang Yuan and Junjie Qiu and Junlong Li and Junxiao Song and Kai Dong and Kai Hu and Kaige Gao and Kang Guan and Kexin Huang and Kuai Yu and Lean Wang and Lecong Zhang and Lei Xu and Leyi Xia and Liang Zhao and Litong Wang and Liyue Zhang and Meng Li and Miaojun Wang and Mingchuan Zhang and Minghua Zhang and Minghui Tang and Mingming Li and Ning Tian and Panpan Huang and Peiyi Wang and Peng Zhang and Qiancheng Wang and Qihao Zhu and Qinyu Chen and Qiushi Du and R. J. Chen and R. L. Jin and Ruiqi Ge and Ruisong Zhang and Ruizhe Pan and Runji Wang and Runxin Xu and Ruoyu Zhang and Ruyi Chen and S. S. Li and Shanghao Lu and Shangyan Zhou and Shanhuang Chen and Shaoqing Wu and Shengfeng Ye and Shengfeng Ye and Shirong Ma and Shiyu Wang and Shuang Zhou and Shuiping Yu and Shunfeng Zhou and Shuting Pan and T. Wang and Tao Yun and Tian Pei and Tianyu Sun and W. L. Xiao and Wangding Zeng and Wanjia Zhao and Wei An and Wen Liu and Wenfeng Liang and Wenjun Gao and Wenqin Yu and Wentao Zhang and X. Q. Li and Xiangyue Jin and Xianzu Wang and Xiao Bi and Xiaodong Liu and Xiaohan Wang and Xiaojin Shen and Xiaokang Chen and Xiaokang Zhang and Xiaosha Chen and Xiaotao Nie and Xiaowen Sun and Xiaoxiang Wang and Xin Cheng and Xin Liu and Xin Xie and Xingchao Liu and Xingkai Yu and Xinnan Song and Xinxia Shan and Xinyi Zhou and Xinyu Yang and Xinyuan Li and Xuecheng Su and Xuheng Lin and Y. K. Li and Y. Q. Wang and Y. X. Wei and Y. X. Zhu and Yang Zhang and Yanhong Xu and Yanhong Xu and Yanping Huang and Yao Li and Yao Zhao and Yaofeng Sun and Yaohui Li and Yaohui Wang and Yi Yu and Yi Zheng and Yichao Zhang and Yifan Shi and Yiliang Xiong and Ying He and Ying Tang and Yishi Piao and Yisong Wang and Yixuan Tan and Yiyang Ma and Yiyuan Liu and Yongqiang Guo and Yu Wu and Yuan Ou and Yuchen Zhu and Yuduan Wang and Yue Gong and Yuheng Zou and Yujia He and Yukun Zha and Yunfan Xiong and Yunxian Ma and Yuting Yan and Yuxiang Luo and Yuxiang You and Yuxuan Liu and Yuyang Zhou and Z. F. Wu and Z. Z. Ren and Zehui Ren and Zhangli Sha and Zhe Fu and Zhean Xu and Zhen Huang and Zhen Zhang and Zhenda Xie and Zhengyan Zhang and Zhewen Hao and Zhibin Gou and Zhicheng Ma and Zhigang Yan and Zhihong Shao and Zhipeng Xu and Zhiyu Wu and Zhongyu Zhang and Zhuoshu Li and Zihui Gu and Zijia Zhu and Zijun Liu and Zilin Li and Ziwei Xie and Ziyang Song and Ziyi Gao and Zizheng Pan},
      year={2025},
      eprint={2412.19437},
      archivePrefix={arXiv},
      primaryClass={cs.CL},
      url={https://arxiv.org/abs/2412.19437}, 
}

@misc{mikolov2013efficientestimationwordrepresentations,
      title={Efficient Estimation of Word Representations in Vector Space}, 
      author={Tomas Mikolov and Kai Chen and Greg Corrado and Jeffrey Dean},
      year={2013},
      eprint={1301.3781},
      archivePrefix={arXiv},
      primaryClass={cs.CL},
      url={https://arxiv.org/abs/1301.3781}, 
}

@misc{liu2024probinglanguagemodelspretraining,
      title={Probing Language Models for Pre-training Data Detection}, 
      author={Zhenhua Liu and Tong Zhu and Chuanyuan Tan and Haonan Lu and Bing Liu and Wenliang Chen},
      year={2024},
      eprint={2406.01333},
      archivePrefix={arXiv},
      primaryClass={cs.CL},
      url={https://arxiv.org/abs/2406.01333}, 
}

@misc{groeneveld2024olmoacceleratingsciencelanguage,
      title={{OLMo}: Accelerating the Science of Language Models}, 
      author={Dirk Groeneveld and Iz Beltagy and Pete Walsh and Akshita Bhagia and Rodney Kinney and Oyvind Tafjord and Ananya Harsh Jha and Hamish Ivison and Ian Magnusson and Yizhong Wang and Shane Arora and David Atkinson and Russell Authur and Khyathi Raghavi Chandu and Arman Cohan and Jennifer Dumas and Yanai Elazar and Yuling Gu and Jack Hessel and Tushar Khot and William Merrill and Jacob Morrison and Niklas Muennighoff and Aakanksha Naik and Crystal Nam and Matthew E. Peters and Valentina Pyatkin and Abhilasha Ravichander and Dustin Schwenk and Saurabh Shah and Will Smith and Emma Strubell and Nishant Subramani and Mitchell Wortsman and Pradeep Dasigi and Nathan Lambert and Kyle Richardson and Luke Zettlemoyer and Jesse Dodge and Kyle Lo and Luca Soldaini and Noah A. Smith and Hannaneh Hajishirzi},
      year={2024},
      eprint={2402.00838},
      archivePrefix={arXiv},
      primaryClass={cs.CL},
      url={https://arxiv.org/abs/2402.00838}, 
}

@misc{pawelczyk2023incontext,
      title={In-Context Unlearning: Language Models as Few Shot Unlearners}, 
      author={Martin Pawelczyk and Seth Neel and Himabindu Lakkaraju},
      year={2023},
      eprint={2310.07579},
      archivePrefix={arXiv},
      primaryClass={cs.LG}
}

@misc{hayes2024inexact,
      title={Inexact Unlearning Needs More Careful Evaluations to Avoid a False Sense of Privacy}, 
      author={Jamie Hayes and Ilia Shumailov and Eleni Triantafillou and Amr Khalifa and Nicolas Papernot},
      year={2024},
      eprint={2403.01218},
      archivePrefix={arXiv},
      primaryClass={cs.LG}
}

@misc{duan2024membership,
      title={Do Membership Inference Attacks Work on Large Language Models?}, 
      author={Michael Duan and Anshuman Suri and Niloofar Mireshghallah and Sewon Min and Weijia Shi and Luke Zettlemoyer and Yulia Tsvetkov and Yejin Choi and David Evans and Hannaneh Hajishirzi},
      year={2024},
      eprint={2402.07841},
      archivePrefix={arXiv},
      primaryClass={cs.CL}
}

@inproceedings{li-etal-2023-mope,
    title = "{M}o{P}e: Model Perturbation based Privacy Attacks on Language Models",
    author = "Li, Marvin  and
      Wang, Jason  and
      Wang, Jeffrey  and
      Neel, Seth",
    editor = "Bouamor, Houda  and
      Pino, Juan  and
      Bali, Kalika",
    booktitle = "Proceedings of the 2023 Conference on Empirical Methods in Natural Language Processing",
    month = dec,
    year = "2023",
    address = "Singapore",
    publisher = "Association for Computational Linguistics",
    url = "https://aclanthology.org/2023.emnlp-main.842",
    doi = "10.18653/v1/2023.emnlp-main.842",
    pages = "13647--13660",
}

@misc{carlini2022first,
      title={Membership Inference Attacks From First Principles}, 
      author={Nicholas Carlini and Steve Chien and Milad Nasr and Shuang Song and Andreas Terzis and Florian Tramer},
      year={2022},
      eprint={2112.03570},
      archivePrefix={arXiv},
      primaryClass={cs.CR}
}

@misc{carlini2021extracting,
  title = {Extracting Training Data from Large Language Models},
  author = {Nicholas Carlini and Florian Tram{\`e}r and Eric Wallace and Matthew Jagielski and Ariel Herbert-Voss and Katherine Lee and Adam Roberts and Tom B. Brown and Dawn Xiaodong Song and {\'U}lfar Erlingsson and Alina Oprea and Colin Raffel},
  year = {2020},
  eprint = {2012.07805},
  archivePrefix = {arXiv},
  url = {https://arxiv.org/abs/2012.07805}
}

@misc{shokri2017membership,
      title={Membership Inference Attacks against Machine Learning Models}, 
      author={Reza Shokri and Marco Stronati and Congzheng Song and Vitaly Shmatikov},
      year={2017},
      eprint={1610.05820},
      archivePrefix={arXiv},
      primaryClass={cs.CR}
}

@misc{yeom2018privacy,
      title={Privacy Risk in Machine Learning: Analyzing the Connection to Overfitting}, 
      author={Samuel Yeom and Irene Giacomelli and Matt Fredrikson and Somesh Jha},
      year={2018},
      eprint={1709.01604},
      archivePrefix={arXiv},
      primaryClass={cs.CR}
}

@article{article,
author = {Sag, Matthew},
year = {2019},
month = {01},
pages = {},
title = {The New Legal Landscape for Text Mining and Machine Learning},
journal = {SSRN Electronic Journal},
doi = {10.2139/ssrn.3331606}
}

@misc{zhou2023quantifying,
      title={Quantifying and Analyzing Entity-level Memorization in Large Language Models}, 
      author={Zhenhong Zhou and Jiuyang Xiang and Chaomeng Chen and Sen Su},
      year={2023},
      eprint={2308.15727},
      archivePrefix={arXiv},
      primaryClass={cs.CL}
}

@inproceedings{mattern2023membership,
    title = "Membership Inference Attacks against Language Models via Neighbourhood Comparison",
    author = "Mattern, Justus  and
      Mireshghallah, Fatemehsadat  and
      Jin, Zhijing  and
      Schoelkopf, Bernhard  and
      Sachan, Mrinmaya  and
      Berg-Kirkpatrick, Taylor",
    booktitle = "Findings of the Association for Computational Linguistics: ACL 2023",
    month = jul,
    year = "2023",
    address = "Toronto, Canada",
    publisher = "Association for Computational Linguistics",
    url = "https://aclanthology.org/2023.findings-acl.719",
    doi = "10.18653/v1/2023.findings-acl.719",
    pages = "11330--11343",
}

@misc{biderman2023pythia,
      title={Pythia: A Suite for Analyzing Large Language Models Across Training and Scaling}, 
      author={Stella Biderman and Hailey Schoelkopf and Quentin Anthony and Herbie Bradley and Kyle O'Brien and Eric Hallahan and Mohammad Aflah Khan and Shivanshu Purohit and USVSN Sai Prashanth and Edward Raff and Aviya Skowron and Lintang Sutawika and Oskar van der Wal},
      year={2023},
      eprint={2304.01373},
      archivePrefix={arXiv},
      primaryClass={cs.CL}
}

@misc{gao2020pile,
      title={{The Pile}: An {800GB} Dataset of Diverse Text for Language Modeling}, 
      author={Leo Gao and Stella Biderman and Sid Black and Laurence Golding and Travis Hoppe and Charles Foster and Jason Phang and Horace He and Anish Thite and Noa Nabeshima and Shawn Presser and Connor Leahy},
      year={2020},
      eprint={2101.00027},
      archivePrefix={arXiv},
      primaryClass={cs.CL}
}

@misc{touvron2023llama,
      title={{LLaMA}: Open and Efficient Foundation Language Models}, 
      author={Hugo Touvron and Thibaut Lavril and Gautier Izacard and Xavier Martinet and Marie-Anne Lachaux and Timothée Lacroix and Baptiste Rozière and Naman Goyal and Eric Hambro and Faisal Azhar and Aurelien Rodriguez and Armand Joulin and Edouard Grave and Guillaume Lample},
      year={2023},
      eprint={2302.13971},
      archivePrefix={arXiv},
      primaryClass={cs.CL}
}

@article{dl_privacy_survey,
author = {Liu, Bo and Ding, Ming and Shaham, Sina and Rahayu, Wenny and Farokhi, Farhad and Lin, Zihuai},
title = {When Machine Learning Meets Privacy: A Survey and Outlook},
year = {2021},
issue_date = {March 2022},
publisher = {Association for Computing Machinery},
address = {New York, NY, USA},
volume = {54},
number = {2},
issn = {0360-0300},
url = {https://doi.org/10.1145/3436755},
doi = {10.1145/3436755},
journal = {ACM Comput. Surv.},
month = {mar},
articleno = {31},
numpages = {36}
}

@misc{kurmanji2023unbounded,
      title={Towards Unbounded Machine Unlearning}, 
      author={Meghdad Kurmanji and Peter Triantafillou and Jamie Hayes and Eleni Triantafillou},
      year={2023},
      eprint={2302.09880},
      archivePrefix={arXiv},
      primaryClass={cs.LG}
}

@misc{shi2023detecting,
title={Detecting Pretraining Data from Large Language Models},
author={Weijia Shi and Anirudh Ajith and Mengzhou Xia and Yangsibo Huang and Daogao Liu and Terra Blevins and Danqi Chen
and Luke Zettlemoyer},
year={2023},
eprint={2310.16789},
archivePrefix={arXiv},
primaryClass={cs.CL}
}

@misc{das2025blindbaselinesbeatmembership,
      title={Blind Baselines Beat Membership Inference Attacks for Foundation Models}, 
      author={Debeshee Das and Jie Zhang and Florian Tramèr},
      year={2025},
      eprint={2406.16201},
      archivePrefix={arXiv},
      primaryClass={cs.CR},
      url={https://arxiv.org/abs/2406.16201}, 
}

@misc{maini2024reassessingemnlp2024s,
  author = {Maini, Pratyush and Suri, Anshuman},
  title = {Reassessing {EMNLP} 2024’s Best Paper: Does Divergence-Based Calibration for Membership Inference Attacks Hold Up?},
  year = {2024},
  note = {Accessed May 1, 2025},
  url  = {https://www.anshumansuri.com/blog/2024/calibrated-mia/}
}

@misc{devlin2019bertpretrainingdeepbidirectional,
      title={{BERT}: Pre-training of Deep Bidirectional Transformers for Language Understanding}, 
      author={Jacob Devlin and Ming-Wei Chang and Kenton Lee and Kristina Toutanova},
      year={2019},
      eprint={1810.04805},
      archivePrefix={arXiv},
      primaryClass={cs.CL},
      url={https://arxiv.org/abs/1810.04805}, 
}

@misc{zhang2025minkimprovedbaselinedetecting,
      title={Min-K\%++: Improved Baseline for Detecting Pre-Training Data from Large Language Models}, 
      author={Jingyang Zhang and Jingwei Sun and Eric Yeats and Yang Ouyang and Martin Kuo and Jianyi Zhang and Hao Frank Yang and Hai Li},
      year={2025},
      eprint={2404.02936},
      archivePrefix={arXiv},
      primaryClass={cs.CL},
      url={https://arxiv.org/abs/2404.02936}, 
}

@misc{dolma,
      title={Dolma: an Open Corpus of Three Trillion Tokens for Language Model Pretraining Research}, 
      author={Luca Soldaini and Rodney Kinney and Akshita Bhagia and Dustin Schwenk and David Atkinson and Russell Authur and Ben Bogin and Khyathi Chandu and Jennifer Dumas and Yanai Elazar and Valentin Hofmann and Ananya Harsh Jha and Sachin Kumar and Li Lucy and Xinxi Lyu and Nathan Lambert and Ian Magnusson and Jacob Morrison and Niklas Muennighoff and Aakanksha Naik and Crystal Nam and Matthew E. Peters and Abhilasha Ravichander and Kyle Richardson and Zejiang Shen and Emma Strubell and Nishant Subramani and Oyvind Tafjord and Pete Walsh and Luke Zettlemoyer and Noah A. Smith and Hannaneh Hajishirzi and Iz Beltagy and Dirk Groeneveld and Jesse Dodge and Kyle Lo},
      year={2024},
      eprint={2402.00159},
      archivePrefix={arXiv},
      primaryClass={cs.CL},
      url={https://arxiv.org/abs/2402.00159}, 
}

@misc{kim2025detectingtrainingdatalarge,
      title={Detecting Training Data of Large Language Models via Expectation Maximization}, 
      author={Gyuwan Kim and Yang Li and Evangelia Spiliopoulou and Jie Ma and Miguel Ballesteros and William Yang Wang},
      year={2025},
      eprint={2410.07582},
      archivePrefix={arXiv},
      primaryClass={cs.CL},
      url={https://arxiv.org/abs/2410.07582}, 
}

@misc{hayes2026exploringlimitsstrongmembership,
      title={Exploring the limits of strong membership inference attacks on large language models}, 
      author={Jamie Hayes and Ilia Shumailov and Christopher A. Choquette-Choo and Matthew Jagielski and George Kaissis and Milad Nasr and Sahra Ghalebikesabi and Meenatchi Sundaram Mutu Selva Annamalai and Niloofar Mireshghallah and Igor Shilov and Matthieu Meeus and Yves-Alexandre de Montjoye and Katherine Lee and Franziska Boenisch and Adam Dziedzic and A. Feder Cooper},
      year={2026},
      eprint={2505.18773},
      archivePrefix={arXiv},
      primaryClass={cs.CR},
      url={https://arxiv.org/abs/2505.18773}, 
}

@misc{meeus2025sokmembershipinferenceattacks,
      title={SoK: Membership Inference Attacks on LLMs are Rushing Nowhere (and How to Fix It)}, 
      author={Matthieu Meeus and Igor Shilov and Shubham Jain and Manuel Faysse and Marek Rei and Yves-Alexandre de Montjoye},
      year={2025},
      eprint={2406.17975},
      archivePrefix={arXiv},
      primaryClass={cs.CL},
      url={https://arxiv.org/abs/2406.17975}, 
}

\appendix
\section{Attack Details.}

\label{apdx:supclas}

\paragraph{Supervised Learning for Classifiers.} We use random forest, logistic regression, and a neural network as supervised classifiers for our model-free blind baselines (traditional word embedding features). For random forest, we use $100$ trees with no limit on maximum depth and the Gini impurity splitting criterion. For logistic regression, we use a default lbfgs solver with $1{,}000$ iteration maximum and L2-regularization. For neural network, we train for $10$ epochs on a batch size of $128$ using the Adam optimizer with a learning rate of $0.001$, with a ReLU architecture of $4$ layers going from input dimension to $250$, to $100$, to $10$, then to $1$ dimensions.

\paragraph{Evaluation Configuration.} We evaluate all attacks using $1{,}000$ train and validation points. For all supervised methods used in the Pythia blind baselines, we train on $4{,}000$ points; for all supervised methods used in OLMo baselines, we train on $8{,}000$ points. 

\paragraph{Other Evaluation Details.} MoPe uses $10$ perturbations with $\sigma = 0.005$, as recommended in the paper~\citep{li-etal-2023-mope}. For Min-K and Min-K++, we use $k=0.1$. For ReCaLL, we use $100$-token long prefixes as the extra conditioning.

\section{Dataset Details}

\paragraph{Pythia Data Construction.} We evaluate a Pythia checkpoint at $97{,}000$ steps into training. To construct the dataset for evaluation, we randomly sample data points from steps $0$ to $97{,}000$, and then from $97{,}000$ to $98{,}500$ (approximately the end of Epoch 1). We run all attacks on Pythia models trained using \emph{deduplicated} training data. 

\paragraph{OLMo Data Construction.} OLMo $7$B was trained on $1.25$ epochs from the $2$T token training corpus Dolma for a total of $556{,}000$ training steps. The remaining $0.25$ epochs after the first epoch are taken from another shuffling of the training corpus. Because our checkpoint-based method is only valid if the model did not see the data after the checkpoint, we restrict our attention to the model state through the first epoch, after $452{,}000$ training steps. We then choose to evaluate MIAs for the model at checkpoint $400{,}000$. We choose the member data by randomly sampling from data that the model saw between checkpoints $0$ and $400{,}000$, and the non-member data by randomly sampling from data the model saw between checkpoints $401{,}000$ and $452{,}000$. Because the entire Dolma dataset already undergoes several different kinds of deduplicating before being used to train OLMo (see Section 5.4 of~\citep{dolma} for details), this guarantees that member and non-member data have the same marginal distributions.  

\section{OLMo Results} 

\label{apdx:olmo}

\paragraph{Blind MIAs.} As with the Pythia models in Section~\ref{sec:results} of the paper, we run blind supervised baselines for the OLMo model as well. These results are given in Table~\ref{tab:olmobaselines}. 

\begin{table}[h!]
\centering
\caption{Sanity check: supervised model-free (``blind") MIAs get no signal on our cleaned split of Dolma.}
\label{tab:olmobaselines}
\begin{tabular}{rcccc}
\toprule
\textbf{Blind MIA} & \textbf{AUC} & \textbf{AUC SE} & $\text{\textbf{TPR}}_{1\%}$ \\ \midrule
BoW (Tokens) & 0.491 & 0.0131 & 0.01 \\
TFIDF (Tokens) & 0.491 & 0.0128 & 0.008 \\
W2V (Text) & 0.494 & 0.0129 & 0.012 \\
BERT (Text) & 0.496 & 0.0126 & 0.009 \\
\bottomrule
\end{tabular}
\end{table}

\paragraph{Other Attacks.} We benchmark various MIAs from previous works, as in Section~\ref{sec:results}, this time against OLMo $7$B. See Table~\ref{tab:olmomias} for full results.


\begin{table}[h!]
\centering
\caption{We benchmark several published MIAs against OLMo $7$B on our cleaned split of Dolma.}
\label{tab:olmomias}
\begin{tabular}{rcccc}
\toprule
\textbf{MIA} & \textbf{AUC} & \textbf{AUC SE} & $\text{\textbf{TPR}}_{1\%}$ \\ \midrule

LOSS & 0.505 & 0.0125 & 0.01 \\
Min-K & 0.504 & 0.0129 & 0.013 \\
Min-K++ & 0.490 & 0.0124 & 0.009 \\
Zlib & 0.499 & 0.0128 & 0.019 \\
ReCaLL & 0.521 & 0.0130 & 0.011 \\

\bottomrule
\end{tabular}
\end{table}


\section{Additional Details}

\paragraph{Compute Estimates.} To run the experiments, we used a compute node with an NVIDIA A$100$ 80GB GPU. All experiments in this paper can be run on a single one of these GPUs. All results for pretrained MIAs are on model sizes $70$M, $160$M, $410$M, $1$B, and $2.8$B for Pythia, and $7$B for OLMo. To run blind baselines, we create features using $4{,}000$ randomly sampled member/non-member points, train a classifier, and evaluate the classifier on $1{,}000$ distinct member/non-member points. Most of these steps are runnable on a consumer laptop. As noted previously, in the pretrained setting, we evaluate all MIAs on $1{\small,}000$ points from member and non-member splits, which requires only running inference on models. In MoPe, we run inference on ten times as many points (because we use ten perturbed models). In total these attacks took around $3$ A$100$ GPU-days. 

\paragraph{AI Assistants.}  While all work was done and checked by the authors, language models were used in the process to refine ideas, write small snippets of code, and tune writing for clarity. 


\end{document}